\documentclass{article}
\usepackage{rat}

\usepackage{times}

\usepackage{soul}
\usepackage{url}
\usepackage[hidelinks]{hyperref}
\usepackage[utf8]{inputenc}
\usepackage[small]{caption}
\usepackage{graphicx}
\usepackage{amsmath}
\usepackage{booktabs}
\usepackage{amssymb}
\usepackage{subcaption}
\usepackage{adjustbox}
\usepackage{multirow}
\usepackage{multicol}
\graphicspath{ {./fig/} }
\usepackage{bbding} 
\usepackage{hyperref}

\title{ Recurrent Affine Transformation for Text-to-image Synthesis}

\author{
Senmao Ye$^1$
\and
Fei Liu$^1$\and
Minkui Tan$^{1}$
\affiliations
$^1$South China University of Technology
\emails
senmaoy@gmail.com,
feiliu@scut.edu.cn,
minkuitan@scut.edu.cn
}
\begin{document}
	
	\maketitle
	
	\begin{abstract}
			Text-to-image synthesis aims to generate natural images conditioned on text descriptions. The main difficulty of this task lies in effectively fusing text information into the image synthesis process. Existing methods usually adaptively fuse suitable text information into the synthesis process with multiple isolated fusion blocks (e.g., Conditional 
			Batch Normalization and Instance Normalization). However, isolated fusion blocks not only conflict with each other but also increase the difficulty of training (see first page of the supplementary). To address these issues, we propose a Recurrent Affine Transformation (RAT) for Generative Adversarial Networks that connects all the fusion blocks with a recurrent neural network to model their long-term dependency. Besides, to improve semantic consistency between texts and synthesized images, we incorporate a spatial attention model in the discriminator. Being aware of matching image regions, text descriptions supervise the generator to synthesize more relevant image contents. Extensive experiments on the CUB, Oxford-102 and COCO datasets demonstrate the superiority of the proposed model in comparison to state-of-the-art models \footnote{https://github.com/senmaoy/Recurrent-Affine-Transformation-for-Text-to-image-Synthesis.git}
		
			. 
	\end{abstract}
	
	\section{Introduction}
	Synthesizing images conditioned on descriptive sentences is a popular research topic in the cross-field of vision and language processing. Due to its various potential applications such as photo-editing, computer-aided design and virtual scene generation, many methods have been proposed to address this problem such as Generative Adversarial Networks (GANs) \cite{DBLP:conf/icml/ReedAYLSL16,DBLP:journals/pami/ZhangXLZWHM19,DBLP:conf/ijcai/FangXCTZ19} and variational auto-encoders~\cite{DBLP:conf/icml/GregorDGRW15,mansimov2015generating}. Recently, GAN-based methods achieve promising results on this task~\cite{DBLP:conf/cvpr/QiaoZXT19,DBLP:journals/corr/abs-2008-05865,DBLP:journals/corr/abs-2104-00567,DBLP:conf/cvpr/ZhuP0019,DBLP:journals/corr/abs-2108-12141}.
	
	GANs usually adaptively fuse suitable text information into the synthesis process with multiple isolated fusion blocks such as Conditional Batch Normalization (CBN) and Instance Normalization (CIN). CIN $et\ al$~\cite{DBLP:conf/iclr/DumoulinSK17} is first proposed for style transfer. Afterwards, BigGAN~\cite{DBLP:conf/iclr/BrockDS19} and Style-GAN~\cite{DBLP:journals/pami/KarrasLA21} synthesize natural images with impressing visual quality based on CBN and CIN, respectively. Recently, 	DF-GAN~\cite{DBLP:journals/corr/abs-2008-05865}, DT-GAN~\cite{DBLP:conf/ijcnn/ZhangS21} and SSGAN~\cite{DBLP:journals/corr/abs-2104-00567} use CIN and CBN to fuse text information into synthesized images.
    Despite their popularity, CIN and CBN suffer from a serious drawback in that they are isolated in different layers, which ignores the global assignment of text information fused in different layers. Furthermore, isolated fusion blocks are hard to optimize since they do not interact with each other.

    In this paper, we propose a Recurrent Affine Transformation (RAT) for controlling all the fusion blocks consistently. RAT  expresses different layers' outputs with standard context vectors of the same shape to achieve unified control of different layers. The context vectors are then connected using Recurrent Neural Network (RNN) in order to detect long-term dependencies. With skipping connections in RNN, fusion blocks are not only consistent between neighbouring blocks but also reduce training difficulty.

Besides, to improve semantic consistency between texts and synthesized images, we incorporate a spatial attention model in the discriminator.	Being aware of matching image regions, text descriptions supervise the generator to synthesize more relevant image contents. Nevertheless, based on extensive experiments, we find that vanilla spatial attention with softmax function leads to model collapse since softmax suppresses most probabilities to be near zero, which in turn leads to increased instability of GAN. To overcome this, we use soft threshold function in place of softmax function, which prevents small attention probabilities from being close to zero. With spatial attention, our discriminator can focus its attention on image regions that are pertinent to the text description, which allows it to supervise the generator more effectively.
		
	The contributions of this paper are the following:
	\begin{itemize}
		\setlength{\topsep}{0pt}
		\setlength{\itemsep}{0pt}
		\setlength{\parsep}{0pt}
		\setlength{\parskip}{0pt}
		
		\item We propose a recurrent affine transformation that connects all the fusion blocks to allow global assignment of text information in the synthesis process.
		\item We incorporate spatial attention into the discriminator to focus on relevant image regions, so the generated images are more relevant to the text description.
		\item It is evident that the proposed model improves the visual quality and evaluation metrics on the CUB, Oxford-102, and COCO datasets.
	\end{itemize}
	
	\section{Related Work}
	\paragraph{Text-to-image Synthesis.}
    Text-to-image synthesis is one of the tasks within conditional image synthesis. With the advent of GANs, conditional GANs have achieved excellent performance on this task. Conditional GAN~\cite{DBLP:journals/corr/MirzaO14} first proposed the conditional version of GAN which directly concatenates noise vector and conditional feature vector. However, text-to-image models based on this work fuse  text information roughly by concatenating the text feature with noise vector. Vincent $et\ al$~\cite{DBLP:conf/iclr/DumoulinSK17} proposed a more advanced fusion method (i.e., CIN) which uses adaptive mean and variance to control the image style.  CIN and its variant are commonly used in recent works. For example, BigGAN~\cite{DBLP:conf/iclr/BrockDS19} and Style-GAN~\cite{DBLP:journals/pami/KarrasLA21} successively use CBN and CIN to achieve impressing visual quality on ImageNet.
    
    Recent developments have utilized CBN and CIN for integrating text information into synthesis.  The Semantic-Spatial aware version of CBN has been proposed by SSGAN~\cite{DBLP:journals/corr/abs-2104-00567} as a way of making CBN aware of spatial regions relevant to text description.  
   To better fuse text information into the synthesis process,
    DF-GAN~\cite{DBLP:journals/corr/abs-2008-05865} proposed a deep fusion method that have multiple affine layers in a single block. Different from previous work, DF-GAN discards normalization operation without decrease in performance, which reduces computational occupation and  limitation from large batch-size. Similarly to DF-GAN, our RAT also discards normalization operations since normalization has no influence on performance.

	Some previous work \cite{DBLP:journals/pami/ZhangXLZWHM19,DBLP:conf/icml/ReedAYLSL16} showed that extra spatial structure information can lead to a great improvement in image quality. The method in \cite{DBLP:conf/icml/ReedAYLSL16} uses key points and bounding boxes to determine what and where to draw in the image. They represented the key points and bounding boxes as a feature map to feed them into the generator. But labelling key points and bound boxes need a lot of human labour and key points are very subjective. Although they also tried to generate key points with an extra GAN, key points are still needed during training.  
	Similar to \cite{DBLP:conf/icml/ReedAYLSL16}, Wang $et\ al$. \cite{wang2016generative} generated images conditioned on surface maps which can reflect rough sketches of the original images. They also use an extra GAN to synthesize surface maps. The above two models learn to generate handcraft annotations so that they can synthesize images from scratch with an explicit notion of image spatial structure. But they both need additional manual annotations to train their model.

	\paragraph{Matching Text and Image Feature.}
	Graves $et\ al$.~\cite{graves2014neural} first proposed an attention model in the context of handwriting synthesis. 	
	Another attention model was proposed by \cite{DBLP:journals/corr/BahdanauCB14} for machine translation. 
	It also uses a weighted summation of local annotations as a context feature. But they computed the weights by softmax function. Following \cite{DBLP:journals/corr/BahdanauCB14}, \cite{DBLP:conf/icml/XuBKCCSZB15} proposed a soft attention model and a hard attention model for image captioning. The soft attention model uses a weighted summation of local region features as a context vector. The hard attention model uses a single local feature as a context feature. The context feature captures visual information relevant to words in the caption. 	Our attention model aims to find image regions relevant to the entire caption. Different from \cite{DBLP:journals/corr/MansimovPBS15}, our model does not directly use attention to generate images. We only apply attention in the discriminator to facilitate the comparison between images and captions.

	\section{Method}
\begin{figure*}[t]
		\centering
\begin{minipage}{1\textwidth}
	\includegraphics[width=\textwidth]{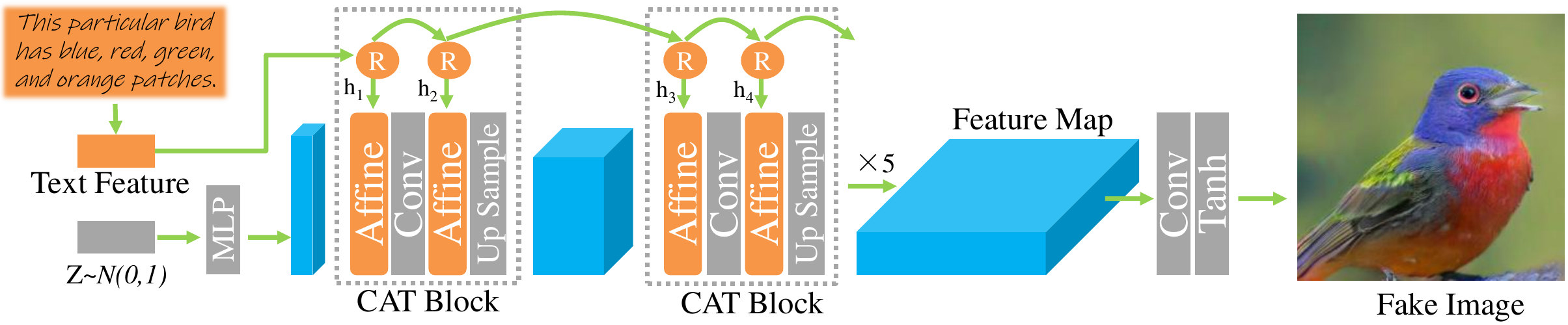}
		\vspace{-22.5pt}
		\caption*{(a)}
\end{minipage}

\begin{minipage}{1\textwidth}
	\includegraphics[width=\textwidth]{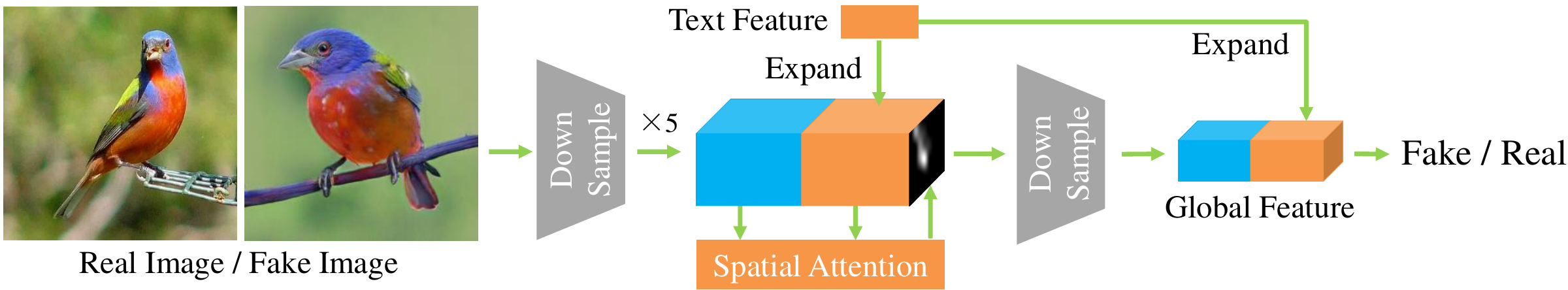}
			\vspace{-19pt}
		\caption*{(b)}
\end{minipage}
		\vspace{-11pt}
	\caption{\label{Fig:generator}(a) Generator with the proposed recurrent affine transformation for text-to-image synthesis. The fusion blocks are connected by a RNN in order to ensure global assignment of text information. (b) Discriminator with spatial attention can focus on image regions relevant to the text description, which helps to judge the authenticity of the input image.}
		\label{Fig:framework}
	\end{figure*}

	 In this section, we aim to construct a GAN that can better control image details with the proposed RAT  (see Figure~\ref{Fig:framework}). We begin with an introduction to the pre-training of text encoder in section~\ref{pre-training}.  We then discuss how to build a generator with recurrent affine transformations in section~\ref{generator}. The final sectionn~\ref{discriminator} describes how to build a discriminator based on spatial attention.
	
	\subsection{Contrastive Text Embedding Pre-training \label{pre-training}}
	
	We use a bidirectional long-short-term-memory network (LSTM) to encode each text description into a sentence-level feature vector ${s}\in \mathbb{R}^{d}$ and a convolution network to encode each image into an image-level feature vector $ {f} \in \mathbb{R}^{d} $. To train these two feature extractors, we adopt a contrastive loss that maximizes the image-text similarity among a batch of training samples.
	Following AttnGAN~\cite{DBLP:conf/cvpr/XuZHZGH018}, we first calculate the similarity matrix for all possible text-image pairs by
	\begin{equation}
		{M}=\left[s_1, s_2,\ldots,s_n\right]^{\mathrm{T}}\left[f_1,f_2,\ldots,f_n\right],
	\end{equation}
	where $n$ is the image number in a batch. $ s_i$ and $f_i $ denote the $i^{th}$ text feature and image feature, respectively. $M_{i,j}$ denotes the dot-product similarity between the $i^{th}$ text feature and $j^{th}$ image feature. The similarity matrix $M$ is turned into match probabilities as follows:
	
	\begin{equation}
		\hat{{M}}_{i, j}=\frac{\exp \left({M}_{i, j}\right)}{\sum_{j=1}^{n} \exp \left({M}_{i, j}\right)} .
	\end{equation}
	To maximize the similarity of text and image features belonging to the same pair, we minimize the contrastive loss:
	\begin{equation}
		\mathcal{L}=-\sum_{i=1}^{n}\log \hat{{M}}_{i, i}.
	\end{equation}
	The pre-trained text feature extractor maps texts into text feature vectors, which reduces the training difficulty of conditional GANs.

	\subsection{ Recurrent Affine Transformation\label{generator}}

	In this section, we aim to use recurrent affine transformation in the generator to enhance the consistency between fusion blocks of different layers. We begin by introducing the proposed  RAT.

    RAT first conducts channel-wise scaling operation on an image feature vector $c$ with the scaling parameter, then it applies the channel-wise shifting operation on $c$ with the shifting parameter. This process can be formally expressed as follows:
	\begin{equation}
		{\rm Affine }\left(c \mid h_t\right)=\gamma_{i} \cdot c+\beta_{i},
	\end{equation}
	where $h_t$ is the hidden state of a RNN and $\gamma,\beta$ are parameters predicted by two one-hidden-layer MLPs conditioned on $h_t$.
	\begin{equation}
		\gamma={\rm MLP}_{1}(h_t), \quad \beta={\rm MLP}_{2}(h_t) .
	\end{equation}When applied to an image feature map composed of $w\times h$ feature vectors, the same affine transformation is repeated for every feature vector. For increasing network depth and non-linearity, as depicted in Figure~\ref{Fig:framework}, several RNN units, RATs and convolutions are stacked to form a RAT block.
	\paragraph{Recurrent Controller.} We use a RNN to model the temporal structure within the RAT blocks in order to assign text information on a global basis. Instead of a vanilla RNN, we use the widely used  LSTM variant. The initial states of the LSTM is computed from the noise vector $z$: 
	\begin{equation}
		h_0={\rm MLP}_{3}(z), \quad c_0={\rm MLP}_{4}(z) .
	\end{equation}
	The update rule of RAT is formulated as follows:
	\begin{eqnarray}
		\left(\begin{array}{cccc}	    \textbf{i}_t  \\ \textbf{f}_t \\	    \textbf{o}_t \\        	{u}_t	\end{array}\right)  &=&
		\left(\begin{array}{cccc}	    \sigma  \\ \sigma \\	    \sigma \\        {\rm tanh}	\end{array}\right)\left( T
		\left(\begin{array}{cccc}	    {s}  \\  \\	    {h}_{t-1} 	\end{array}\right) \right),\\
		\textbf{c}_t &=& \textbf{f}_t\odot \textbf{c}_{t-1} + \textbf{i}_t \odot {u}_t, \\
		{h}_t &=& \textbf{o}_t\odot {\rm tanh}(\textbf{c}_t),\\
		\gamma_t ,\beta_t &=& { }{\rm MLP_1^t}({h}_t), { }{\rm MLP_2^t}({h}_t),
	\end{eqnarray}
	where $\textbf{i}_t,\textbf{f}_t,\textbf{o}_t$ are the input gate, forget gate and output gate, respectively. $T : R^{D+d} \rightarrow R^d$ is an affine transformation, where $ d $ is the dimensionality of the text embedding and $ D $ is the quantity of the RNN hidden state units. 
	
	Furthermore, two layer-specific MLPs are used to convert each hidden state $h_t$ into $\gamma_t $ and $\beta_t$ because each convolution layer has different channel dimensions. Different to conditional instance normalization, conditional batch normalization and conditional deep fusion~\cite{DBLP:journals/corr/abs-2008-05865}, our work no longer takes affine transformation as isolated modules. In contrast, we employ RNN to model the long-term dependency between fusion blocks, which not only forces the fusion blocks to be consistent with one another but also reduces the difficulty of training with skip connections.
	
Finally, as depicted in Figure~\ref{Fig:framework} (a), we use a one-stage generator consisted of 6 up-sample blocks to synthesize fake images. The noise vector $z \sim N(0,1)$ is sampled from standard Gaussian distribution and fed into the generator at the beginning. Each up-sample block is followed by a RAT block to control the image contents. The last $256\times256$ feature map is turned into a fake image of size $256\times256\times 3$ by a hyperbolic tangent function.
	
	\subsection{Matching-Aware Discriminator with Spatial Attention \label{discriminator}}
	To improve the semantic consistency between synthesized image and text description, we incorporate spatial attention into our discriminator.  Being aware of matching image regions, text descriptions supervise the generator to synthesize more relevant image contents.
	As depicted in Figure ~\ref{Fig:framework} (b), several down-sample blocks are used to encode the image into an image feature map $P$. Combing the information in the image feature map $P$ and sentence vector $s$, spatial attention produces an attention map ${\alpha}$ that suppresses  sentence vectors for irrelevant regions.	The precise formulation of this attention model is:
	
	\begin{eqnarray}
		x_{w,h} &=& {\rm MLP}(P_{w,h},s),\\
		\alpha_{w,h}  &=& \frac{\frac{1}{1+e^{-x_{w,h}}}}{\sum_{w=1,h=1}^{W,H}\frac{1}{1+e^{-x_{w,h}}}}, \\
		S_{w,h}  &=&  s\times \alpha_{w,h},
	\end{eqnarray}
	where $S_{w,h}$ is the feature channels of $ S $ at location $\{w,h\}$. $P_{w,h}$ and $s$ are fed into a multi-layer perception with one hidden layer to compute an energy value $x_{w,h}$, then this energy value is converted into attention probabilities $\alpha_{w,h}$. Finally, $S$ is concatenated with $P$ and fed into the latter down-sample blocks to produce the global feature presentation.	For stabilizing GAN training, attention probabilities $\alpha$ are predicted by soft threshold function 	~\cite{DBLP:journals/tip/YeHL18}:
		\begin{equation}  p(x_k) = \frac{\frac{1}{1+e^{-x_k}}}{\sum_{j=1}^{K} \frac{1}{1+e^{-x_j}}}. \end{equation}
    Soft threshold uses a standard logistic function to squash the negative energy values between $(0,1)$ before normalization.     We do not adopt popular softmax function since it maximizes the biggest probability and suppresses other probabilities to be near 0. The extremely small probabilities hamper the back-propagation of the gradients, which worsens the instability of GAN training. In contrast, the soft threshold function prevents attention probabilities from being close to zero and increases the efficiency of back propagation.
	The spatial attention model assigns more text features to relevant image regions, which helps the discriminator to determine whether the text image pair matches. In adversarial training, the stronger discriminator forces the generator to synthesize more relevant image content.
	
	\subsection{Objective Functions}
	Similar to~\cite{DBLP:conf/icml/ReedAYLSL16}, the training objective for the discriminator takes the synthesized image and the mismatched images as negative samples. To keep gradient smooth, we use hinge loss with MA-GP~\cite{DBLP:journals/corr/abs-2008-05865} on real and match text pair. The precise formulation of the training objective of the discriminator is formulated as follows:
	\begin{equation}
		\begin{aligned}
			\mathcal{L}_{\text {adv }}^{D}=& \mathbb{E}_{x \sim p_{\text {data }}}[\max (0,1-D(x, s))] \\
			&+\frac{1}{2} \mathbb{E}_{x \sim p_{G}}[\max (0,1+D(\hat{x}, s))] \\
			&+\frac{1}{2} \mathbb{E}_{x \sim p_{\text {data }}}[\max (0,1+D(x, \hat{s}))],
		\end{aligned}
	\end{equation}
	where s is the given text description, $\hat{s}$ is a mismatched text description. The corresponding training objective of the generator is :
	\begin{equation}
		\mathcal{L}_{\text {adv }}^{G}=\mathbb{E}_{x \sim p_{G}}[\min(D(x, s))].
	\end{equation}
	
	\section{Experiments}
	
	\paragraph{Datasets.} We report results on the popular CUB, Oxford-102 and MS COCO datasets. The CUB dataset has 200 different categories with 11,788 images of birds in total. The Oxford-102 dataset contains 102 different categories with 8,189 images of flowers in total. As in \cite{reed2016learning,reed2016generative}, we split images into class-disjoint training and test sets. The CUB dataset has 150 train classes and 50 test classes, while the Oxford-102 dataset has 82 train+val and 20 test classes. For both datasets, 10 captions are used per image. The MS COCO dataset totally consists of 123,287 images with 5 sentence annotations. The official training split of COCO is used for training and the official validation split of COCO is used for testing. During mini-batch selection for training, a random image view (e.g. crop, flip) is chosen for one of the captions. 
	
	\paragraph{Training Details.}
	The text encoder is frozen during the training of GAN with an output of size $256$. The random noise is sampled from a 100-dimensional standard Gaussian distribution.
	Adam optimizer is used to optimize the network with base learning rates of 	0.0001 for the generator and 0.0004 for the discriminator. We used a mini-batch size of 24 to train the model for 600 epochs on the CUB and Oxford-102 datasets. The training of the CUB dataset takes around 3 days on a single NVIDIA RTX 3090 Ti GPU. To accelerate training, We used a mini-batch size of 48 and train the model for 300 epochs on the COCO dataset. With two RTX 3090 Ti GPUs, the COCO dataset takes approximately two weeks to train. 
	\begin{figure*}[t!h]
		\centering
		
		\begin{minipage}[c]{0.01\textwidth}
			\fontsize{2.0pt}{0.5\baselineskip}\selectfont \center{\ } 
		\end{minipage}
		\hfill
		\begin{minipage}[t]{0.115\textwidth}
			\center{\scriptsize{A small bird with blue-grey wings, rust colored sides and white collar.}}
		\end{minipage}
		\hfill
		\begin{minipage}[t]{0.115\textwidth}
			\center{\scriptsize{This bird is white from crown to belly, with gray wingbars and retrices.}} 
		\end{minipage}
		\hfill
		\begin{minipage}[t]{0.115\textwidth}
			\center{\scriptsize{This bird is mainly grey, it has brown on the feathers and back of the tail.}}
		\end{minipage}
		\hfill
		\begin{minipage}[t]{0.115\textwidth}
			\center{\scriptsize{A bird with blue head, white belly and breast, and the bill is pointed.}}
		\end{minipage}
		\hspace{1pt}
		\begin{minipage}[t]{0.115\textwidth}
			\center{\scriptsize{This flower has a lot of dark red petals and no visible outer stigma or stamen.}} 
		\end{minipage}
		\hfill
		\begin{minipage}[t]{0.115\textwidth}
			\center{\scriptsize{This flower has petals that are purple and bunched together.}} 
		\end{minipage}
		\hfill
		\begin{minipage}[t]{0.115\textwidth}
			\center{\scriptsize{This flower has large smooth white petals that turn yellow toward the center.}}
		\end{minipage}
		\hfill
		\begin{minipage}[t]{0.115\textwidth}
			\center{\scriptsize{A pale purple five petaled flower with yellow stamen and green stigma.}}
		\end{minipage}
		\vspace{2pt}
		
		\begin{minipage}[c]{0.01\textwidth}
			\center{\rotatebox{90}{GT}}
		\end{minipage}
		\hfill
		\begin{minipage}{0.115\textwidth}
			\includegraphics[width=\textwidth]{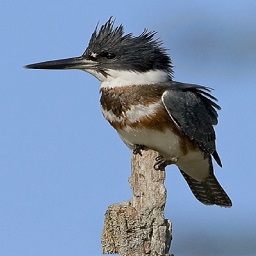}
		\end{minipage}
		\hfill
		\begin{minipage}{0.115\textwidth}
			\includegraphics[width=\textwidth]{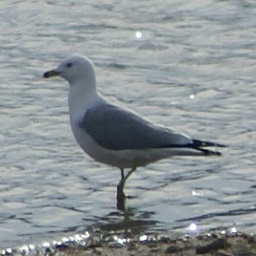}
		\end{minipage}
		\hfill
		\begin{minipage}{0.115\textwidth}
			\includegraphics[width=\textwidth]{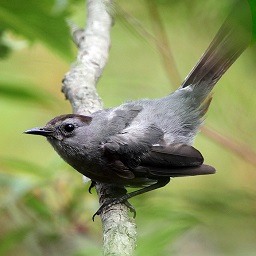}
		\end{minipage}
		\hfill
		\begin{minipage}{0.115\textwidth}
			\includegraphics[width=\textwidth]{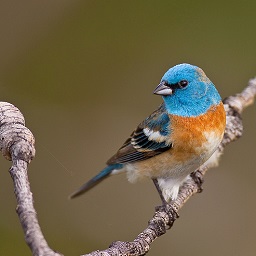}
		\end{minipage}
		\hspace{1pt}
		\begin{minipage}{0.115\textwidth}
			\includegraphics[width=\textwidth]{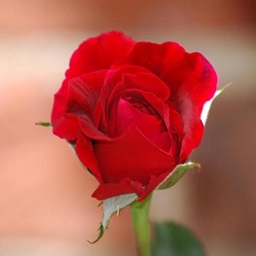}
		\end{minipage}
		\hfill
		\begin{minipage}{0.115\textwidth}
			\includegraphics[width=\textwidth]{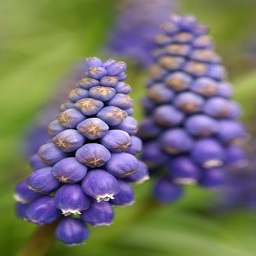}
		\end{minipage}
		\hfill
		\begin{minipage}{0.115\textwidth}
			\includegraphics[width=\textwidth]{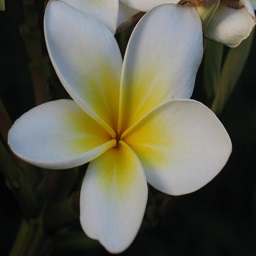}
		\end{minipage}
		\hfill
		\begin{minipage}{0.115\textwidth}
			\includegraphics[width=\textwidth]{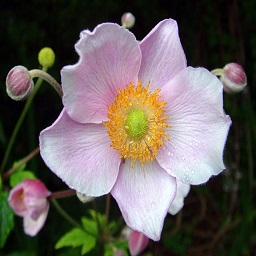}
		\end{minipage}
		\vspace{5pt}
		
		\begin{minipage}[c]{0.01\textwidth}
			\center{\rotatebox{90}{Stack-GAN}}
		\end{minipage}
		\hfill
		\begin{minipage}{0.115\textwidth}
			\includegraphics[width=\textwidth]{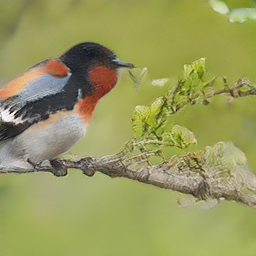}
		\end{minipage}
		\hfill
		\begin{minipage}{0.115\textwidth}
			\includegraphics[width=\textwidth]{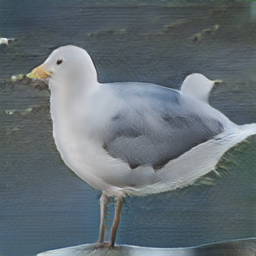}
		\end{minipage}
		\hfill
		\begin{minipage}{0.115\textwidth}
			\includegraphics[width=\textwidth]{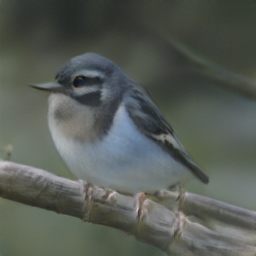}
		\end{minipage}
		\hfill
		\begin{minipage}{0.115\textwidth}
			\includegraphics[width=\textwidth]{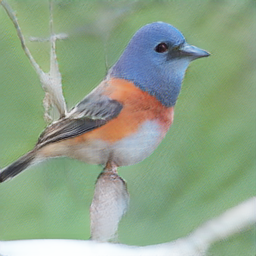}
		\end{minipage}
		\hspace{1pt}
		\begin{minipage}{0.115\textwidth}
			\includegraphics[width=\textwidth]{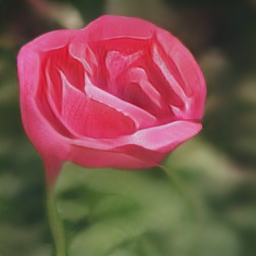}
		\end{minipage}
		\hfill
		\begin{minipage}{0.115\textwidth}
			\includegraphics[width=\textwidth]{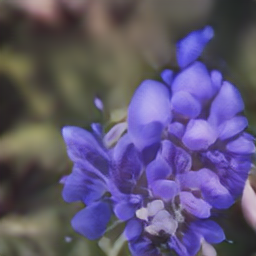}
		\end{minipage}
		\hfill
		\begin{minipage}{0.115\textwidth}
			\includegraphics[width=\textwidth]{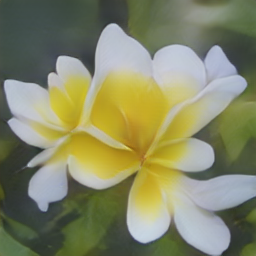}
		\end{minipage}
		\hfill
		\begin{minipage}{0.115\textwidth}
			\includegraphics[width=\textwidth]{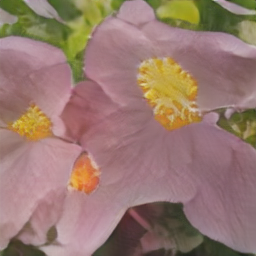}
		\end{minipage}
		\vspace{2pt}
		
		\begin{minipage}[c]{0.01\textwidth}
			\center{\rotatebox{90}{DF-GAN}}
		\end{minipage}
		\hfill
		\begin{minipage}{0.115\textwidth}
			\includegraphics[width=\textwidth]{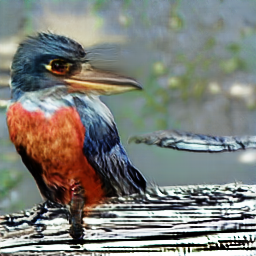}
		\end{minipage}
		\hfill
		\begin{minipage}{0.115\textwidth}
			\includegraphics[width=\textwidth]{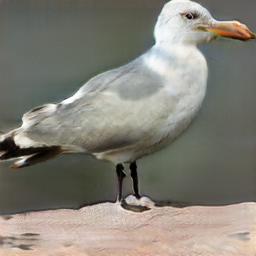}
		\end{minipage}
		\hfill
		\begin{minipage}{0.115\textwidth}
			\includegraphics[width=\textwidth]{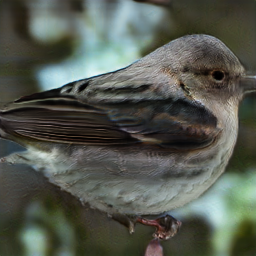}
		\end{minipage}
		\hfill
		\begin{minipage}{0.115\textwidth}
			\includegraphics[width=\textwidth]{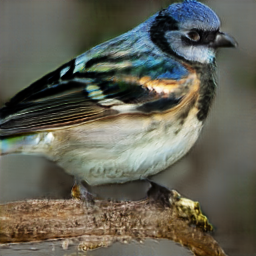}
		\end{minipage}
		\hspace{1pt}
		\begin{minipage}{0.115\textwidth}
			\includegraphics[width=\textwidth]{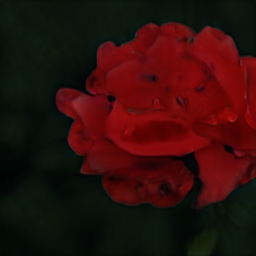}
		\end{minipage}
		\hfill
		\begin{minipage}{0.115\textwidth}
			\includegraphics[width=\textwidth]{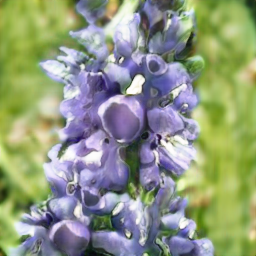}
		\end{minipage}
		\hfill
		\begin{minipage}{0.115\textwidth}
			\includegraphics[width=\textwidth]{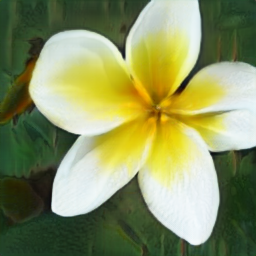}
		\end{minipage}
		\hfill
		\begin{minipage}{0.115\textwidth}
			\includegraphics[width=\textwidth]{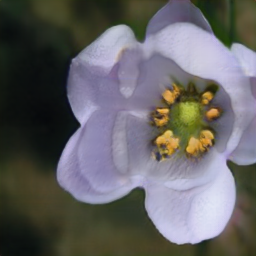}
		\end{minipage}
		\vspace{2pt}
		
		\begin{minipage}[c]{0.01\textwidth}
			\center{\rotatebox{90}{Ours}}
		\end{minipage}
		\hfill
		\begin{minipage}{0.115\textwidth}
			\includegraphics[width=\textwidth]{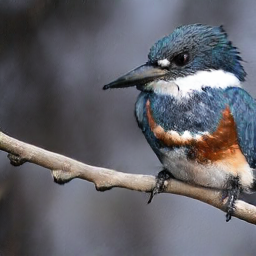}
		\end{minipage}
		\hfill
		\begin{minipage}{0.115\textwidth}
			\includegraphics[width=\textwidth]{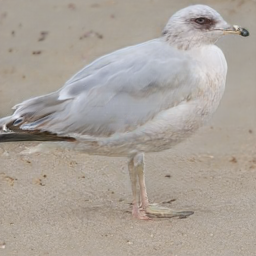}
		\end{minipage}
		\hfill
		\begin{minipage}{0.115\textwidth}
			\includegraphics[width=\textwidth]{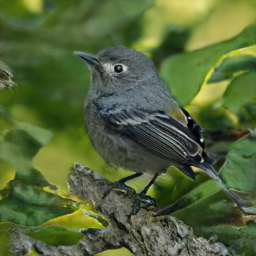}
		\end{minipage}
		\hfill
		\begin{minipage}{0.115\textwidth}
			\includegraphics[width=\textwidth]{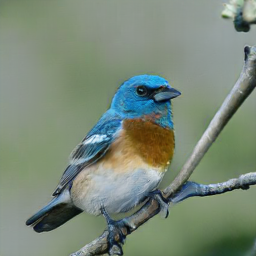}
		\end{minipage}
		\hspace{1pt}
		\begin{minipage}{0.115\textwidth}
			\includegraphics[width=\textwidth]{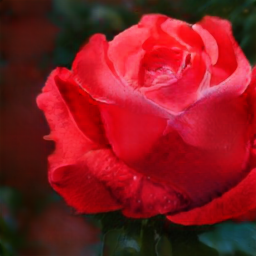}
		\end{minipage}
		\hfill
		\begin{minipage}{0.115\textwidth}
			\includegraphics[width=\textwidth]{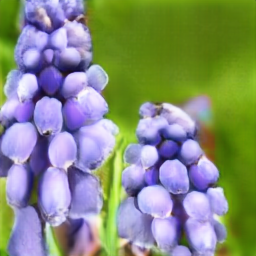}
		\end{minipage}
		\hfill
		\begin{minipage}{0.115\textwidth}
			\includegraphics[width=\textwidth]{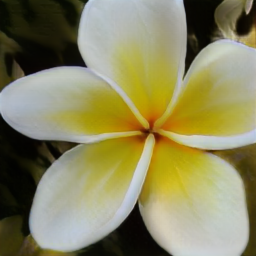}
		\end{minipage}
		\hfill
		\begin{minipage}{0.115\textwidth}
			\includegraphics[width=\textwidth]{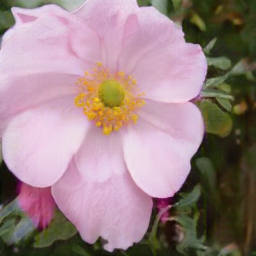}
		\end{minipage}
		\vspace{2pt}
		\caption{Qualitative comparison on the CUB and Ox-ford dataset. The input text descriptions are given in the first row and the corresponding generated images from different methods are shown in the same column. Best view in color and zoom in.}
		\label{fig:qualitative_cub}
		\vspace{-4mm}
	\end{figure*}
	\paragraph{Evaluation Metrics.}
	We adopt the widely used Inception Score (IS)~\cite{DBLP:conf/nips/SalimansGZCRCC16} and Fr\'{e}chet Inception Distance (FID)~\cite{DBLP:conf/nips/HeuselRUNH17} to quantify the quantitative performance. On the MS COCO dataset, an Inception-v3 network pre-trained on Image-net is used to compute the KL-divergence between the conditional class distribution (generated images) and the marginal class distribution (real images). The presence of a large IS indicates that the generated images are of high quality. The FID computes the Fr\'{e}chet Distance between the image features distribution of the generated and real-world images. The image features are extracted by the same pre-trained Inception v3 network. A lower FID implies the generated images are closer to the real images. To evaluate the IS and FID scores, 30k images are generated by each model from the test dataset as previous works~\cite{DBLP:journals/corr/abs-2104-00567,DBLP:journals/corr/abs-2008-05865,DBLP:conf/cvpr/LiZZHHLG19,DBLP:conf/ijcnn/ZhangS21} reported that the IS metric completely fails in evaluating the synthesized images. Therefore, we only compare the FID on the COCO dataset. On the CUB and Oxford-102 dataset, pre-trained Inception models are fine-tuned on two fine-grained classification tasks~\cite{DBLP:journals/pami/ZhangXLZWHM19}.
    
	\paragraph{Compared Models.} We compare our model with recent state-of-the-art methods: StackGAN++~\cite{DBLP:journals/pami/ZhangXLZWHM19}, DM-GAN~\cite{DBLP:conf/cvpr/ZhuP0019}, DF-GAN~\cite{DBLP:journals/corr/abs-2008-05865}, DAE-GAN~\cite{DBLP:journals/corr/abs-2108-12141}, SSACN~\cite{DBLP:journals/corr/abs-2104-00567}, AttnGAN~\cite{DBLP:conf/cvpr/XuZHZGH018},
	DTGAN~\cite{DBLP:conf/ijcnn/ZhangS21}.
	\subsection{Comparisons with Others}
	\paragraph{Quantitative Results.}	For simplicity, we denote GAN with the proposed RAT as RAT-GAN.
	\begin{table}[t!]
		\caption{Performance of IS and FID of StackGAN++, AttnGAN, SSGAN, DM-GAN, DTGAN, DF-GAN and our method on the CUB, Oxford and COCO test set. The results are taken from the authors' own papers.
			The best results are in bold.}
		\label{tab:results}
		\vspace{-2mm}
		\centering
		\begin{adjustbox}{max width=1\textwidth}
			\begin{tabular}{lcccc}
				\toprule
				\multirow{2}*{Methods} &\multicolumn{2}{c}{IS $\uparrow$}& \multicolumn{2}{c}{FID $\downarrow$} \cr 
				\cmidrule(lr){2-3} \cmidrule(lr){4-5} 
				&CUB&Oxford&CUB& COCO\\
				\midrule 
				StackGAN++   & 4.04 $\pm$ .06 &3.26  &{15.30}  &81.59\\
				AttnGAN            & 4.36 $\pm$ .03&-   &23.98  &35.49\\
				DAE-GAN                       & 4.42 $\pm$ .04 &-  & 15.19 & 28.12 \\
				DM-GAN                 & 4.75 $\pm$ .07 &-  &16.09  &32.64\\
				DTGAN            & 4.88 $\pm$ .03  &-  &16.35 &23.61\\
				DF-GAN              & 5.10 $\pm$ .- -  &3.80  &14.81 &21.42\\
				SSAG                                  & {5.17 $\pm$ .08}&-& 15.61 & {19.37}\\
				Ours                                   & \textbf{5.36 $\pm$ .20}&\textbf{4.09}   &\textbf{13.91} & \textbf{14.60}\\
				\bottomrule
			\end{tabular}
		\end{adjustbox}
		\vspace{-2mm}
	\end{table}
	
	In this section, we present results on the CUB dataset of bird images, MS COCO dataset of common object images and Oxford-102 dataset of flower images in Table~\ref{tab:results}. On the CUB dataset, our model achieves the highest Inception scores (5.36) and lowest FID scores (13.91) compared with other state-of-the-art models. On the Oxford dataset, our model achieves the highest Inception scores (4.09). Results of DF-GAN on the Oxford dataset are based on its public available code on the Github.	The superiority of our RAT-GAN is more obvious on the COCO dataset where images are more complicated. According to the results, the proposed recurrent fusion strategy performs better in complex scenes.  Noticeably, our model has no extra supervision such as DAMSM loss (SSGAN) or cycle consisted loss (MirrorGAN). We also do not fine-tune the text encoder or truncate the image feature space or the noise vector space as DF-GAN. Extensive results well prove the effectiveness of the proposed RAT.

\begin{figure}[t!h]
	\centering
	
	\begin{minipage}[c]{0.01\textwidth}
		\fontsize{2.0pt}{0.5\baselineskip}\selectfont \center{\ } 
	\end{minipage}
	\hfill
	\begin{minipage}[t]{0.11\textwidth}
		\center{\scriptsize{A police man on a motorcycle is idle in front of a bush.}}
	\end{minipage}
	\hfill
	\begin{minipage}[t]{0.11\textwidth}
		\center{\scriptsize{A man riding a wave on top of a surfboard.}} 
	\end{minipage}
	\hfill
	\begin{minipage}[t]{0.11\textwidth}
		\center{\scriptsize{Some red and\\ green flower in a room.}}
	\end{minipage}
	\hfill
	\begin{minipage}[t]{0.11\textwidth}
		\center{\scriptsize{Assorted electronic devices sitting together in a photo.}}
	\end{minipage}
	
	\vspace{2pt}
	
		\begin{minipage}[c]{0.01\textwidth}
		\center{\rotatebox{90}{DF-GAN}}
	\end{minipage}
	\hfill
	\begin{minipage}{0.11\textwidth}
		\includegraphics[width=\textwidth]{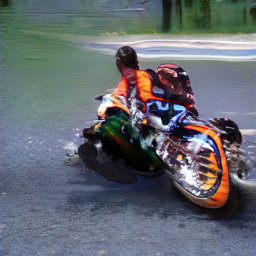}
	\end{minipage}
	\hfill
	\begin{minipage}{0.11\textwidth}
		\includegraphics[width=\textwidth]{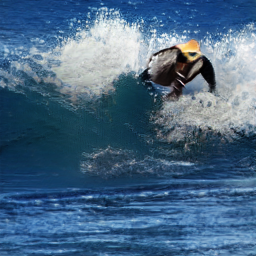}
	\end{minipage}
	\hfill
	\begin{minipage}{0.11\textwidth}
		\includegraphics[width=\textwidth]{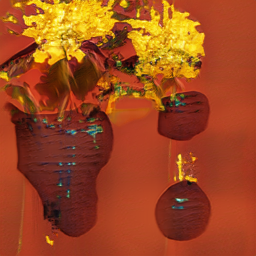}
	\end{minipage}
	\hfill
	\begin{minipage}{0.11\textwidth}
		\includegraphics[width=\textwidth]{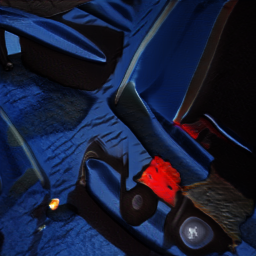}
	\end{minipage}	
	
\vspace{3pt}

	\begin{minipage}[c]{0.01\textwidth}
		\center{\rotatebox{90}{Ours}}
	\end{minipage}
	\hfill
	\begin{minipage}{0.11\textwidth}
		\includegraphics[width=\textwidth]{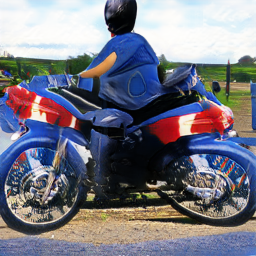}
	\end{minipage}
	\hfill
	\begin{minipage}{0.11\textwidth}
		\includegraphics[width=\textwidth]{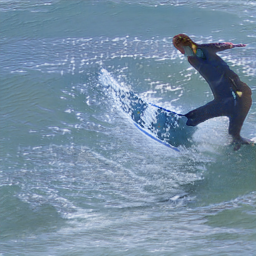}
	\end{minipage}
	\hfill
	\begin{minipage}{0.11\textwidth}
		\includegraphics[width=\textwidth]{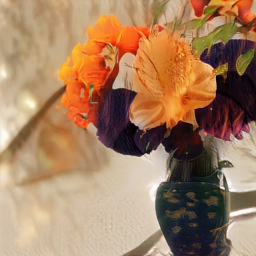}
	\end{minipage}
	\hfill
	\begin{minipage}{0.11\textwidth}
		\includegraphics[width=\textwidth]{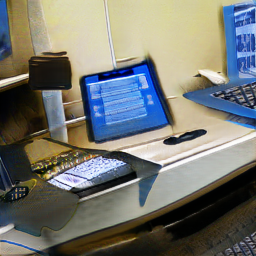}
	\end{minipage}

	\vspace{2pt}

	\caption{Qualitative comparison between DF-GAN and our model  on the COCO dataset. Best view in color and zoom in.}
	\label{Fig:coco}
	\vspace{-4mm}
\end{figure}

	\paragraph {Qualitative Results.}
	In this section, we present qualitative results on the CUB dataset	of bird images and the Oxford-102 dataset of flower images. We compare the visualization results of stackGAN, DF-GAN and the proposed RAT-GAN in Figure ~\ref{fig:qualitative_cub}. StackGAN is a classical multi-stage method and DF-GAN is a popular one-stage method for text-to-image synthesis.
	On the CUB dataset, we observe that our model performs much better than DF-GAN and stackGAN with clear details such as feathers, eyes and feet. What's more, the background is also more regular in the results of our RAT-GAN. On the Oxford dataset, our RAT-GAN has better texture and more relevant colors than others.	 With the proposed RAT and the spatial attention model, our model has fewer distorted shapes and more relevant contents than the other two models. In addition, we observe that stackGAN suppresses the image pixels to be 0.5/1 in $128\times 128$ and $256\times 256$ branches (the first $64\times 64$ branch is normal), which leads to vaguer images than the other two models.
    As depicted in Figure~\ref{Fig:coco}, the qualitative superiority of RAT-GAN is more obvious on the COCO dataset compared with DF-GAN because the recurrent interactions between fusion blocks make them as an unit, which enables RAT to realize more complicated control over synthesized images. More synthesized images are included in the supplementary.
 
	\subsection{Ablation Studies}
	\begin{table}[t!]
		\caption{Ablation study of RAT and spatial attention in our model on the test set of the CUB dataset. \textit{MLPs} means replacing RAT blocks with stacked MLP blocks and \textit{shallow} means removing half of the affine layers from a RAT block.}
		\label{tab:ablation_study_components}
		\vspace{-4mm}
		\begin{center}
			\begin{tabular}{c c c c c}
				\hline
				\multirow{2}{*}{ID} & \multicolumn{2}{c}{Components} & \multirow{2}{*}{IS $\uparrow$} & \multirow{2}{*}{FID $\downarrow$} \\
				\cline{2-3} & RAT & Spatial Att \\
				\hline 
				0 & - & - & 4.86 $\pm$ 0.04 & 19.24 \\
				\hline
				1 & MLPs & - & 4.51 $\pm$ 0.19 & 22.57 \\
				\hline
				2 & \checkmark (shallow) & - & 5.20 $\pm$ 0.11 & 14.90 \\
				\hline   
				3 & \checkmark & - & 5.25 $\pm$ 0.22 & 14.86 \\                     
				\hline
				4 &\checkmark & \checkmark & \textbf{5.36 $\pm$ 0.20} & \textbf{13.91} \\
				\hline
			\end{tabular}
		\end{center}
		\vspace{-4mm}
	\end{table}
	\paragraph{Generator.}For ablation study, we show quantitative evaluation results of model components in Table~\ref{tab:ablation_study_components}. 
	The baseline is a DF-GAN without latent space truncation. ID 1 directly stacks six MLP blocks to substitute RAT blocks, which decreases the performance because stacked MLPs are hard to optimize. Comparing ID 0 and 3, we can see that RAT improves the model performance obviously. In ID 2, fewer affine layers decrease the performance slightly. 
	
\begin{figure}[b]
	\centering
	
	\begin{minipage}[c]{0.01\textwidth}
		\fontsize{2.0pt}{0.5\baselineskip}\selectfont \center{\ } 
	\end{minipage}
	
	\begin{minipage}{0.091\textwidth}
		\includegraphics[width=\textwidth]{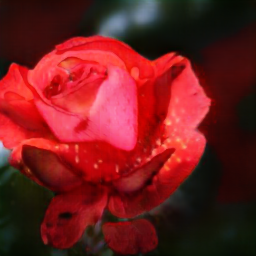}
	\end{minipage}
	\begin{minipage}{0.091\textwidth}
		\includegraphics[width=\textwidth]{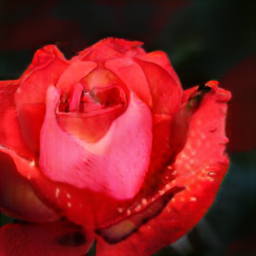}
	\end{minipage}
	\begin{minipage}{0.091\textwidth}
		\includegraphics[width=\textwidth]{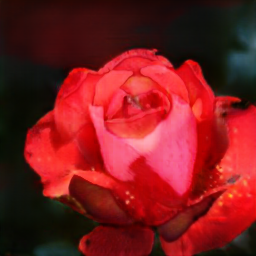}
	\end{minipage}
	\begin{minipage}{0.091\textwidth}
		\includegraphics[width=\textwidth]{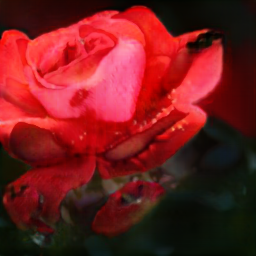}
	\end{minipage}
	\begin{minipage}{0.091\textwidth}
		\includegraphics[width=\textwidth]{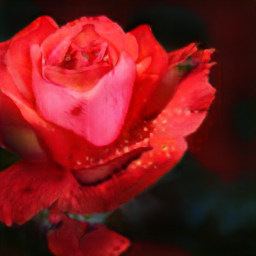}
	\end{minipage}

		\begin{minipage}{0.091\textwidth}
		\includegraphics[width=\textwidth]{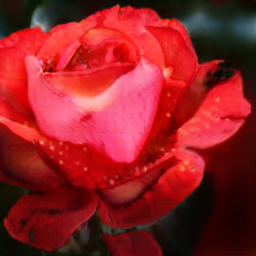}
	\end{minipage}
	\begin{minipage}{0.091\textwidth}
		\includegraphics[width=\textwidth]{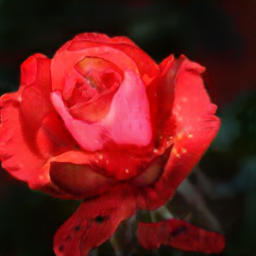}
	\end{minipage}
	\begin{minipage}{0.091\textwidth}
		\includegraphics[width=\textwidth]{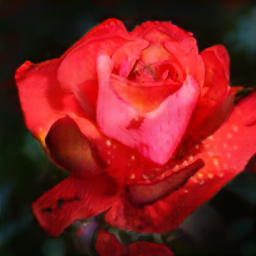}
	\end{minipage}
	\begin{minipage}{0.091\textwidth}
		\includegraphics[width=\textwidth]{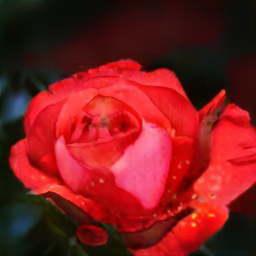}
	\end{minipage}
	\begin{minipage}{0.091\textwidth}
		\includegraphics[width=\textwidth]{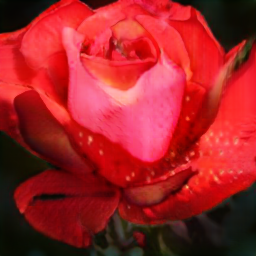}
	\end{minipage}
\begin{minipage}[t]{0.5\textwidth}
		\center{\scriptsize{This flower is dark red in color, with petals that are curled closely around the center. \noindent \noindent \noindent}}
	\end{minipage}

	\caption{Randomly generated images conditioned on the same text description with different noise vectors.}
	\label{Fig:variaty}
\end{figure}
	
	\paragraph{Discriminator.}By equipping the discriminator with spatial attention, the performance increases from 5.25 to 5.36, which reveals that enhancing the discriminator helps the generator to synthesize better images. Spatial attention with softmax function leads to model collapse (the generated pixels are all zero), hence the IS and FID are not available. In contrast, spatial attention with soft threshold avoids such model collapse by protecting small probabilities from being zero.

	\paragraph{Diversity.} To qualitatively evaluate the diversity of the proposed DF-GAN, we generate random images conditioned on the same text description and different noise vectors. In Figure~\ref{Fig:variaty}, we display 10 images from the same text. The images share similar foreground with high diversity in spatial structure, which demonstrates that the RAT-GAN can well control the image contents.
	
	\subsection{Visualization of Attention Maps}

	\begin{figure}[t!]
	\centering
	
	\begin{minipage}[c]{0.01\textwidth}
		\fontsize{2.0pt}{0.5\baselineskip}\selectfont \center{\ } 
	\end{minipage}
	\hfill
	\begin{minipage}[t]{0.11\textwidth}
		\center{\scriptsize{This bird has white, light brown, and medium brown small splotches.}}
	\end{minipage}
	\hfill
	\begin{minipage}[t]{0.11\textwidth}
		\center{\scriptsize{This bird is blue and white in color with a stubby beak, and black eye rings.}} 
	\end{minipage}
	\hfill
	\begin{minipage}[t]{0.11\textwidth}
		\center{\scriptsize{This flower is rose shaped with orange overlapping petals and green sepals.}}
	\end{minipage}
	\hfill
	\begin{minipage}[t]{0.11\textwidth}
		\center{\scriptsize{The petals of the flower are purple in color and have green leaves.}}
	\end{minipage}

	\vspace{2pt}
	
	\begin{minipage}[c]{0.01\textwidth}
		\center{\rotatebox{90}{Images}}
	\end{minipage}
	\hfill
	\begin{minipage}{0.11\textwidth}
		\includegraphics[width=\textwidth]{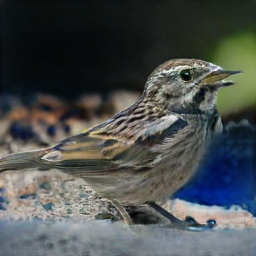}
	\end{minipage}
	\hfill
	\begin{minipage}{0.11\textwidth}
		\includegraphics[width=\textwidth]{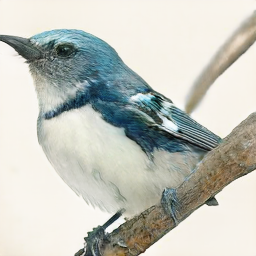}
	\end{minipage}
	\hfill
	\begin{minipage}{0.11\textwidth}
		\includegraphics[width=\textwidth]{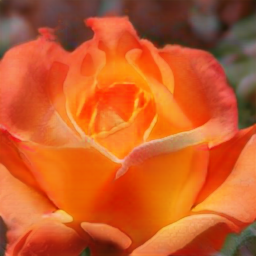}
	\end{minipage}
	\hfill
	\begin{minipage}{0.11\textwidth}
		\includegraphics[width=\textwidth]{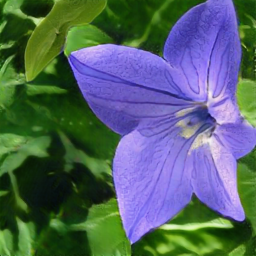}
	\end{minipage}

	\vspace{3pt}
	
	\begin{minipage}[c]{0.01\textwidth}
		\center{\rotatebox{90}{Masks}}
	\end{minipage}
	\hfill
	\begin{minipage}{0.11\textwidth}
		\includegraphics[width=\textwidth]{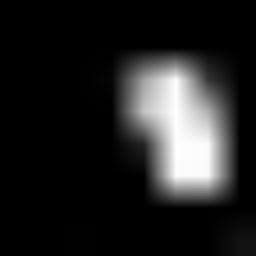}
	\end{minipage}
	\hfill
	\begin{minipage}{0.11\textwidth}
		\includegraphics[width=\textwidth]{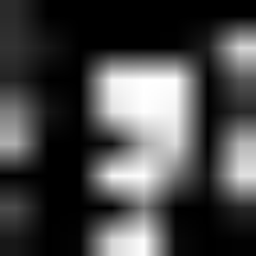}
	\end{minipage}
	\hfill
	\begin{minipage}{0.11\textwidth}
		\includegraphics[width=\textwidth]{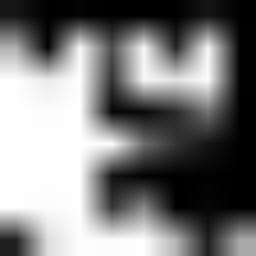}
	\end{minipage}
	\hfill
	\begin{minipage}{0.11\textwidth}
		\includegraphics[width=\textwidth]{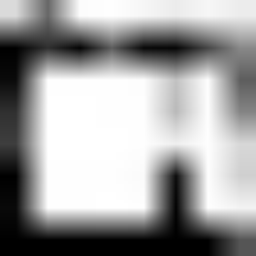}
	\end{minipage}
	
	\vspace{2pt}

	\caption{Attention maps predicted by the spatial attention model. Best view in color and zoom in.}
	\label{Fig:att_mask}
	\vspace{-4mm}
\end{figure}

	To verify the effectiveness of our spatial attention model, we visualize the attention map $\alpha$. 
	The attention map is of size $8\times 8$ with attention probabilities distributed in (0,1). For better visualize the attention probabilities in the discriminator, we simply up-sample the weights by a factor of 32 with bi-linear interpolation and normalize the attention map by: \begin{equation}\alpha' = \frac{(\alpha-min(\alpha))}{(max(\alpha)-min(\alpha))}.\end{equation}
	Some attention maps are visualized in Figure ~\ref{Fig:att_mask}. It's evident that spatial attention can identify regions relevant to the caption, thus enabling the discriminator to make a better comparison between the image and caption.

	\section{Conclusion and Future work}
	In this paper, 	we address the text-to-image problem by GAN with the proposed recurrent affine transformation.  The main difficulty of this task lies in effectively fusing text information into the image synthesis process. Previous models usually use isolated fusion blocks to adaptively fuse suitable information. In contrast,	RAT successfully improves image quality by adding interactions among fusion blocks through RNN. The mutual interactions not only ensure consistent between neighbouring blocks but also
reduce training difficulty.
	Besides, to improve semantic consistency, we incorporate a spatial attention model into the discriminator. Extensive experiments on different datasets demonstrate that our model improves the image quality obviously. In the future, it's interesting to apply RAT to other conditional image synthesis tasks.
	\bibliographystyle{named}
	\bibliography{rat}
			\begin{figure*}[t!hb]

		\begin{subfigure}[t]{0.5\linewidth}
			\centering
			\captionsetup{justification=centering}
			\includegraphics[width=3.6in]{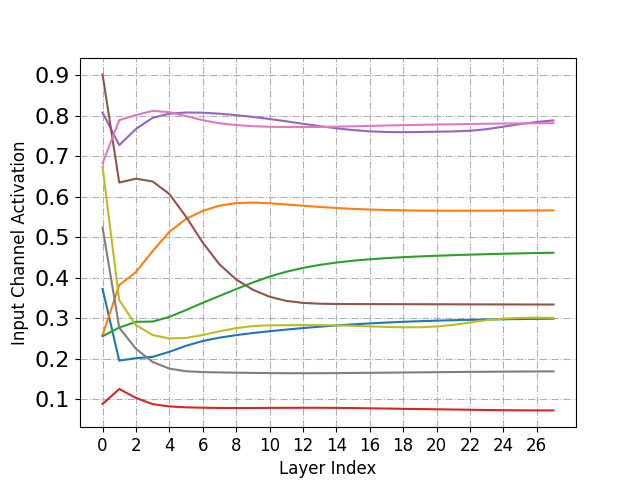}
			\caption{Input activation variation	}
			\label{fig:side:a}
		\end{subfigure}%
		\begin{subfigure}[t]{0.5\linewidth}
			\centering
			\captionsetup{justification=centering}
			\includegraphics[width=3.6in]{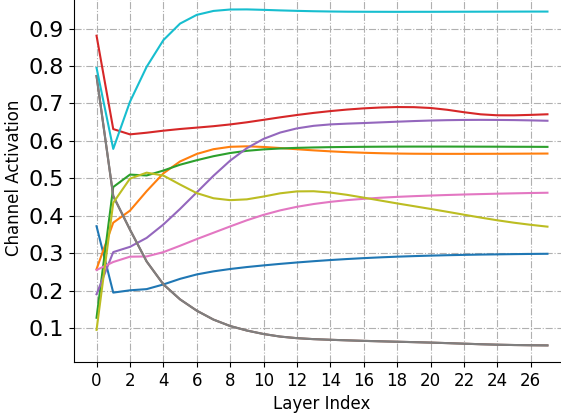}
			\caption{Output activation variation
			}
			\label{fig:side:b}

		\end{subfigure}

			\begin{subfigure}[t]{0.5\linewidth}
		\centering
		\captionsetup{justification=centering}
		\includegraphics[width=3.6in]{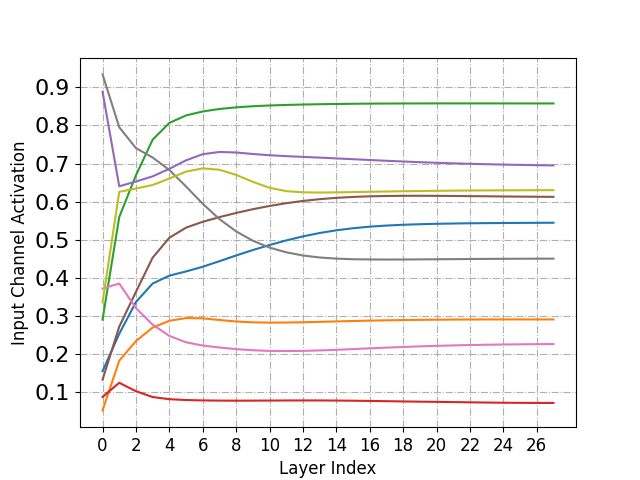}
		\caption{Input activation variation	}
		\label{fig:side:a}
	\end{subfigure}%
	\begin{subfigure}[t]{0.5\linewidth}
		\centering
		\captionsetup{justification=centering}
		\includegraphics[width=3.6in]{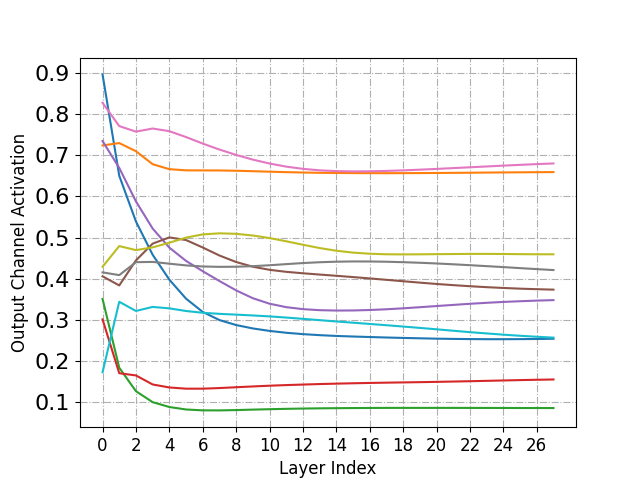}
		\caption{Output activation variation
		}
		\label{fig:side:b}

	\end{subfigure}
	
			\begin{subfigure}[t]{0.5\linewidth}
		\centering
		\captionsetup{justification=centering}
		\includegraphics[width=3.6in]{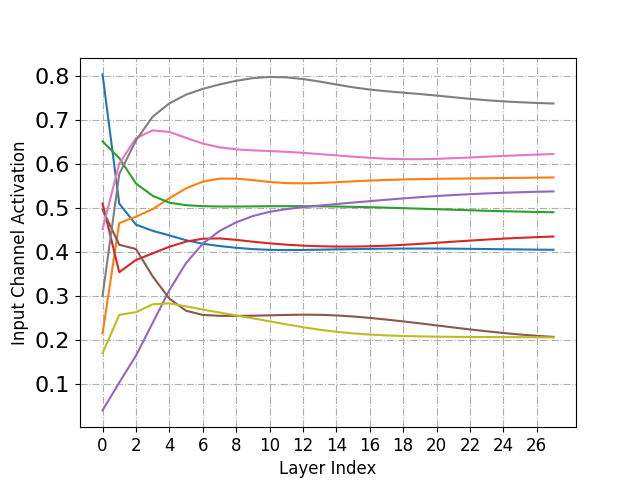}
		\caption{Input activation variation	}
		\label{fig:side:a}
	\end{subfigure}%
	\begin{subfigure}[t]{0.5\linewidth}
		\centering
		\captionsetup{justification=centering}
		\includegraphics[width=3.6in]{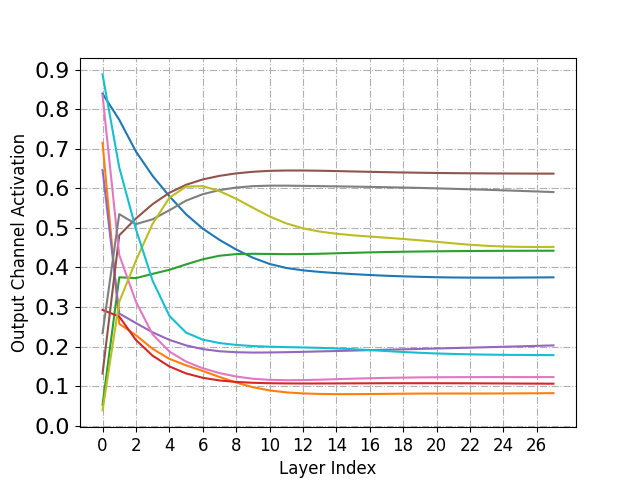}
		\caption{Output activation variation
		}
		\label{fig:side:b}

	\end{subfigure}
		
		\caption{ Variation of channel activations in different layers (28 in total). Probabilities in  input gate $I_t$ and output gate $O_t$ are regarded as the activations of corresponding channels.  Each color stands for a random selected channel. We display 30 input gates and 30 output gate in total.	It's obvious that each channel of $h_t$ are preferred by different layers. What's more, the variation of preference between each layer is smooth, which proves that consistency is naturally required by neighbouring layers
		}
	\end{figure*}

		\begin{figure*}[t]
		\centering
		\includegraphics[height = 6.3cm]{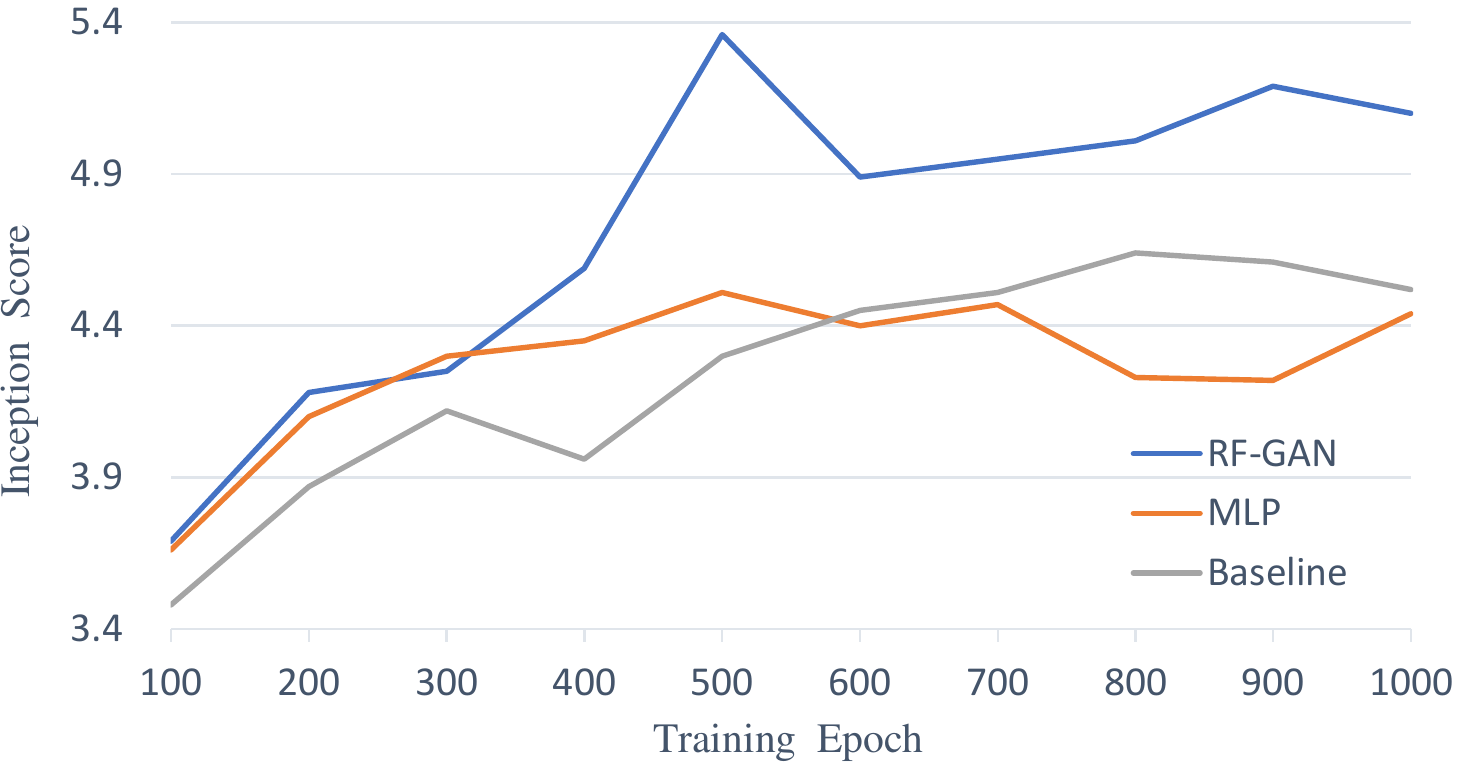}
		\caption{\label{Fig:generator}The variation of Inception Scores during training on the CUB dataset. The proposed RAT-GAN, MLP and the baseline are compared. We observe that connecting CAT blocks with RNN (RAT-GAN) or MLP blocks (MLP) leads to faster  convergence than the baseline because isolated fusion blocks are connected as an unit. But stacked MLP blocks are hard to optimize, hence MLP soon saturates.}
		\label{Fig:loss}
	\end{figure*}	

		\begin{figure*}[t]
	\centering
	\includegraphics[height = 10.3cm]{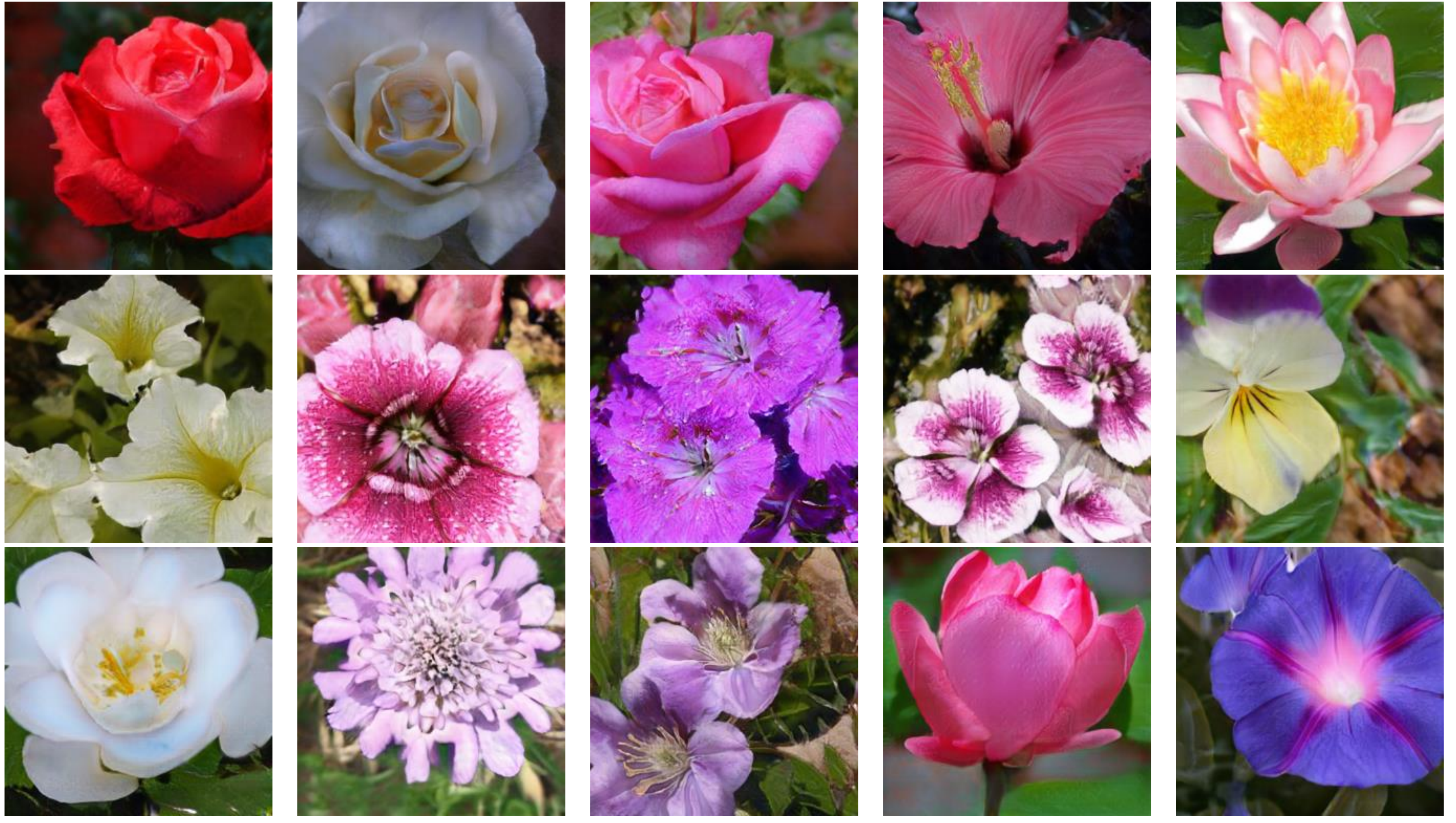}
	\caption{\label{Fig:generator}More visualization samples from  the Oxford dataset.}
	\label{Fig:loss}
\end{figure*}	

		\begin{figure*}[t]
	\centering
	\includegraphics[height = 22.3cm]{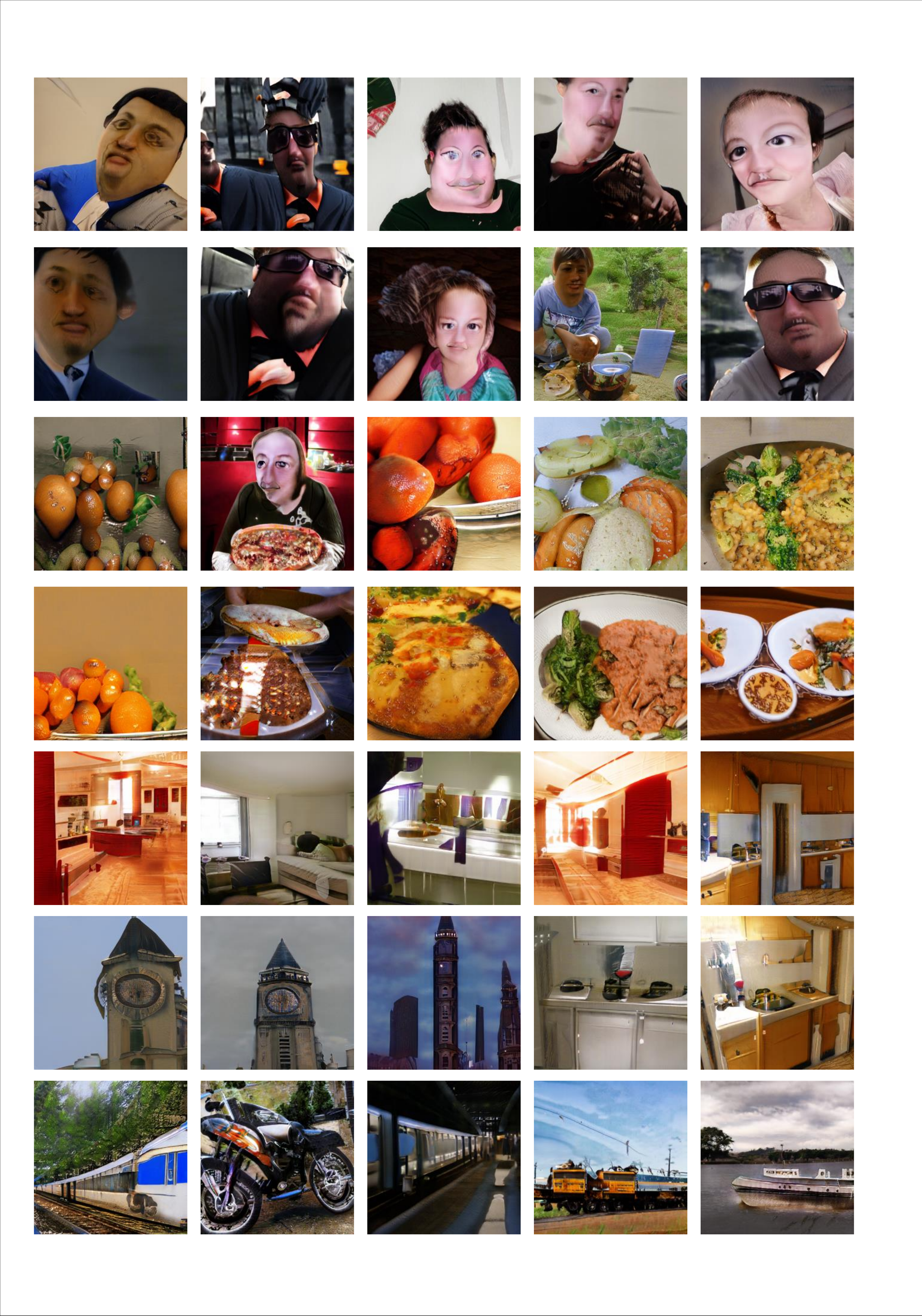}
	\caption{\label{Fig:generator}More visualization samples from the COCO dataset.}
	\label{Fig:loss}
\end{figure*}

\end{document}